\documentclass[letterpaper]{article}
\usepackage{aaai}
\usepackage{times}
\usepackage{helvet}
\usepackage{courier}

\usepackage{amsmath, amssymb}
\usepackage{latexsym}
\usepackage{hyperref}
\usepackage{url}
\usepackage{xcolor}
\usepackage{bm}
\usepackage{multirow}
\usepackage{framed}
\usepackage{array}
\usepackage{wrapfig}
\usepackage{ulem}
\usepackage{color}
\usepackage{graphicx}

\definecolor{newcolor}{rgb}{.8,.349,.1}

\DeclareMathOperator*{\argmin}{arg\,min} % Jan Hlavacek
\begin{document}
\title{Towards Interpretable Deep Neural Networks by Leveraging \\ Adversarial Examples}
% \author{Yinpeng Dong}
% \author{Fan Bao}
% \author{Hang Su}
% \author{Jun Zhu\corref{cor1}}
% \cortext[cor1]{Corresponding author:
%   e-mail: dcszj@mail.tsinghua.edu.cn}
\author{Yinpeng Dong \and Fan Bao \and Hang Su \and Jun Zhu\\
Tsinghua National Lab for Information Science and Technology\\
State Key Lab of Intelligent Technology and Systems\\
Center for Bio-Inspired Computing Research\\
Department of Computer Science and Technology, Tsinghua University, Beijing 100084, China
}
\maketitle
% \begin{abstract}
% \begin{quote}
% Deep neural networks (DNNs) have demonstrated impressive performance on a wide array of tasks, while are usually considered opaque since their internal structures and learned parameters are not interpretable. Ambiguity, a phenomenon that a neuron may fire for various unrelated concepts, hinders the interpretability of DNNs. In this paper, we examine the internal representations and ambiguity of DNNs using adversarial examples, which enables us to analyze the interpretability of DNNs in the perspective of their failures. We find that the learned features of DNNs are inconsistent with human-understandable semantic concepts, especially on adversarial image subset, making it problematic for interpreting and understanding the representations inside DNNs. Based on this analysis, we further propose an adversarial training algorithm by introducing a consistent loss to suppress the ambiguity of neurons, such that the neurons are more consistent with human-interpretable concepts and the network is more interpretable. Extensive experiments show that the features of a DNN trained by our method are more consistent with semantic concepts than those trained normally with the same architecture. The induced interpretable representations enable us to trace eventual outcomes back to influential neurons. Therefore, human users can know how the models make predictions as well as errors.
% \end{quote}
% \end{abstract}

\section{Introduction}
% Deep neural networks (DNNs) have demonstrated unprecedented performance improvements in various computer vision tasks\cite{DeepReview_Lecun_2015}, including image classification\cite{krizhevsky2012imagenet,he2015deep}, object detection\cite{girshick2014rich,He2017Mask}, etc.
% However, the composite non-linear structure makes DNNs highly non-transparent, which are therefore treated as ``black-box'' models due to the lack of understandable decoupled components and the unclear working mechanism\cite{bengio2013representation}.
% In some cases, it limits the applicability of DNNs when the models are incapable of explaining the reasons behind the decisions or actions to human users, since it is far from enough to provide eventual outcomes to users especially in mission-critical applications, e.g., healthcare or autonomous driving.
% Users need to understand the rationale of the decisions such that they can understand, validate, edit, trust a learned model, and fix the potential problems when it fails or makes errors.

Sometimes it is not enough for a DNN to produce an outcome. For example, in applications such as healthcare, users need to understand the rationale of the decisions. Therefore, it is imperative to develop algorithms to learn models with good interpretability~\cite{DoshiKim2017Interpretability}. % making users clearly understand, appropriately trust, and effectively interact with the models\cite{DoshiKim2017Interpretability}.
An important factor that leads to the lack of interpretability of DNNs is the ambiguity of neurons, where a neuron may fire for various unrelated concepts. %As illustrated in ~\autoref{fig:overview}, a neuron fires for a set of natural images with concept \textit{parrot}, but if we evaluate on some special points of images, e.g., adversarial images, this neuron will fire for images with other concepts which are far from \textit{parrot}. 
%If we view neurons as concept detectors\cite{erhan2009visualizing,zhou2014object,simonyan2013deep,zeiler2014visualizing,springenberg2014striving}, then ambiguity phenomena suggests that these detectors are deficient. These detectors will fail on certain subset of image space(e.g., adversarial image subset), hindering the interpretability of DNNs on the whole image space.
This work aims to increase the interpretability of DNNs on the whole image space by reducing the ambiguity of neurons. In this paper, we make the following contributions:%, which is an important approach towards interpretable DNNs.% The challenge to solve this problem is that the complex structure of networks makes it infeasible to control the semantic concept of each neuron of DNNs. Besides, it is also difficult to diagnose how reliable these neurons are as detectors.

\begin{itemize}
\item We propose a metric to evaluate the consistency level of neurons in a network quantitatively.
\item We find that the learned features of neurons are ambiguous by leveraging adversarial examples.
\item We propose to improve the consistency of neurons on adversarial example subset by an adversarial training algorithm with a consistent loss. %The training procedure encourages the  accordance between the learned features of neurons and semantic concepts.
% Using this metric, we compare the internal interpretability of normally trained models and adversarially trained models.
% \item 
% We present an exemplar-based visualization of predictions. The visualization provides a cue to explain the rationale of the model's predictions. Humans can also know how the model is fooled by adversarial samples by leveraging the visualization.
\end{itemize}

\begin{figure}[!t]
  \centering
    \includegraphics[width=0.9\linewidth]{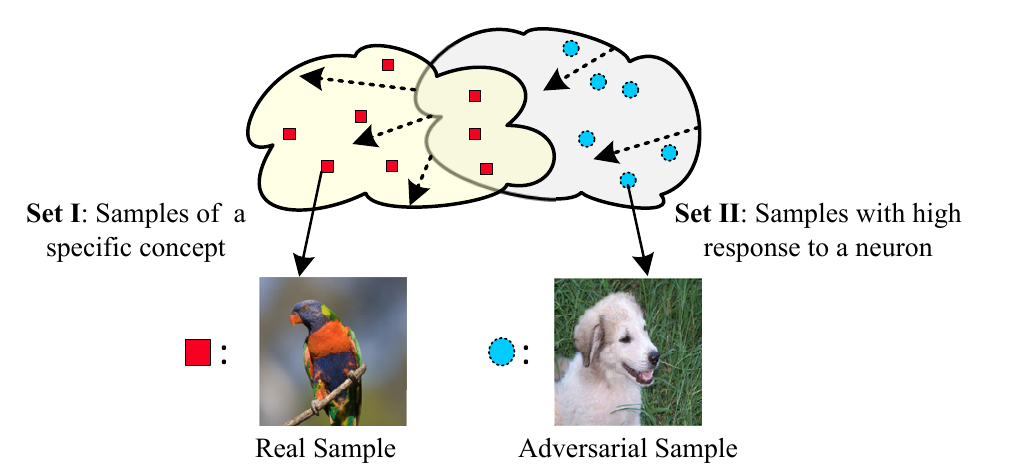}
\caption{Demonstration of the ambiguity of a neuron. When showing real images only, a neuron admittedly detects a semantic meaningful concept, but there also exist samples (e.g., adversarial samples marked with blue circles) in the high dimensional space that are different from the semantic concept but cause high response of the neuron. To reduce the ambiguity of a neuron, we propose to train DNNs adversarially by introducing a consistent loss, such that the neurons are encouraged to respond to related similar concepts or a sole concept (shown as the dashed lines).}
    \label{fig:overview}
\end{figure}
\section{Ambiguity of neurons}
As illustrated in \autoref{fig:overview}, a neuron fires for a set of natural images with concept \textit{parrot}, but if we evaluate on some special points of images, e.g., adversarial images, this neuron will fire for images with other concepts which are far from \textit{parrot}. We call the property that a neuron fires for various unrelated concepts \textbf{ambiguity} of a neuron and these special points of images \textbf{singular points}. On the other hand, if we can hardly find singular points for a neuron, we say that the neuron is \textbf{consistent} and call this property \textbf{consistency} of a neuron. %Although singular points may distribute sparsely in the image space, we can't deny that the ambiguity property of neurons highly reduces the interpretability of DNNs and is an important factor that causes the black-box nature of DNNs.

\subsection{Metric}
\label{sec:consis_metric}
\begin{figure}[!t]
  \centering
    \includegraphics[width=0.5\textwidth]{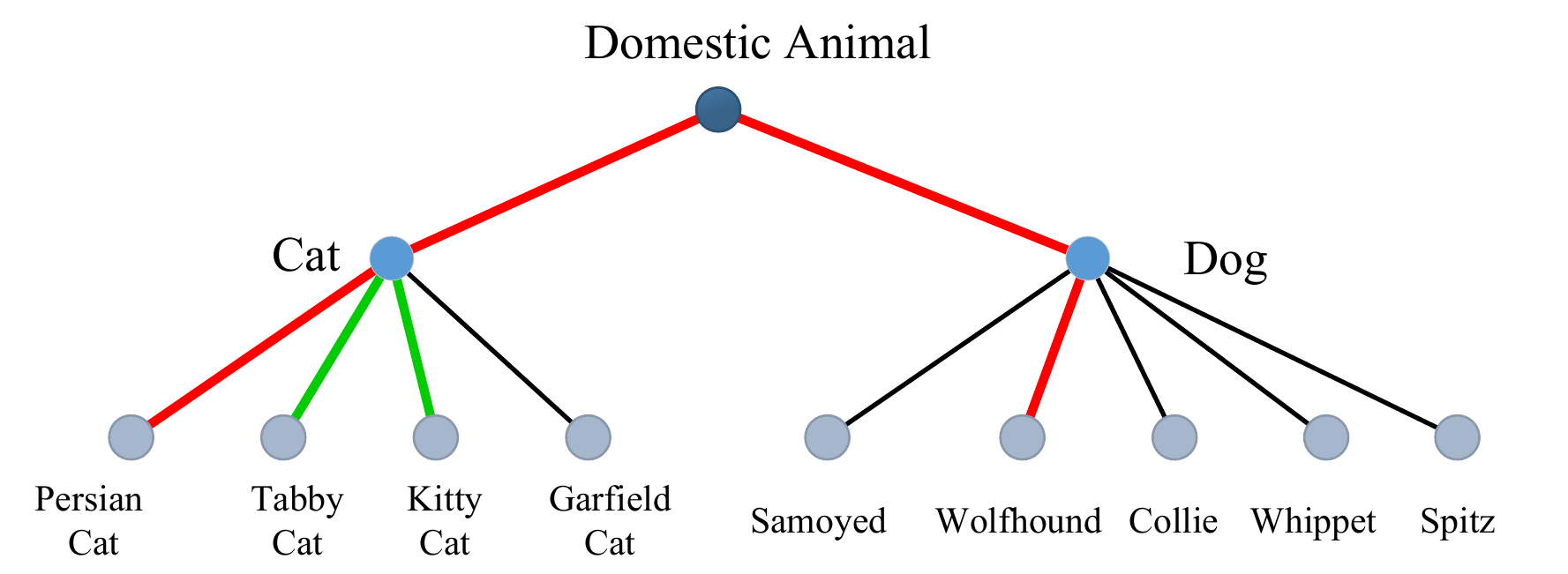}
    \caption{Illustration for quantifying the distance of different classes based on WordNet. The red path indicates the distance between \textit{persian cat} and \textit{wolfhound} $d=4$, which is longer than the distance between \textit{tabby cat} and \textit{kitty cat} ($d=2$).}
    \label{fig:WordNet}
\end{figure}

In this section, we introduce the metric to evaluate the consistency of neurons.
%We first reemphasize the definition of singular points, ambiguity and consistency:
% \begin{myDef}
% A singular point for a neuron is the kind of image which can activate this neuron but doesn't align with the main concept of this neuron.
% \end{myDef}
% \begin{myDef}
% Ambiguity is a property of a neuron that it fires for singular points.
% \end{myDef}
% \begin{myDef}
% Consistency is a property of a neuron that singular points of this neuron can hardly be found (i.e. singular points only take a tiny proportion).
% \end{myDef}
We first define the consistency level between a neuron $n$ and a concept $c$:
\begin{equation}
\label{eq:consis}
\begin{aligned}
consis(n, c)&=Pr(x\,contains\,c|x\,activates\,n)
\end{aligned}
\end{equation}
where $Pr$ is the probability measure of image space.
The consistency metric above is related to a certain concept. To evaluate the consistency of a neuron itself, we need a concept independent metric. 
%Entropy might be a choice at first glance. But a concept $c$ might occur in different images, resulting in $\sum_{c}consis(n, c)$ not necessarily equal to $1$, so entropy can't be used here. Besides, different concepts are not totally unrelated. For example, a neuron which fires for \textit{tabby cat} and \textit{tiger cat} is more consistent than a neuron which fires for \textit{tabby cat} and \textit{plane}. 
Considering correlation between concepts , we quantify the consistency of neurons based on WordNet~\cite{miller1990introduction}.
As demonstrated in \autoref{fig:WordNet}, we measure the distance between different classes on the WordNet tree. We define the correlation between the corresponding concepts as
\begin{equation}
a_{i,j} = \exp\left(-\frac{d^2(w_i, w_j)}{2\sigma^2}\right),
\end{equation}
where $w_i$, $w_j$ are the words of the $i, j$-th classes, $d(w_i, w_j)$ is their WordNet tree distance, and $\sigma$ is a hyper-parameter to control the decaying rate.
We form the distance matrix by collecting each pair of the corresponding classes as $\mathbf{A}=[a_{i,j}]$. 
We further collect $p_i=consis(n, c_i)$ for all concepts $c_i$ into a vector $p$ and the consistency level of a neuron $n$ is quantified as following
\begin{equation}\label{equ_INPscore}
\begin{aligned}
consis(n) &= \|\bm{p}\|_{\mathbf{A}}^2 = \bm{p}^T\mathbf{A}\bm{p}\\
\end{aligned}
\end{equation}
Similarly, we also measure the consistency level constrained on the adversarial samples, by replacing $Pr$ with $Pr_{adv}$, where $Pr_{adv}(\cdot)=Pr(\cdot|x\,is\,an\,adversarial\,sample)$. %Although the adversarial samples may only account for a small proportion of the whole image space, it is still worth considering them separately because they are the subset where the networks are easiest to fail. We denote the consistency level constrained on adversarial samples as $consis_{adv}(n)$.

\section{Methodology}
\label{sec:methodology}
In this section, we introduce the methods. We first illustrate the inconsistency between the learned features of DNNs and semantic meaningful concepts, which motivates us to train DNNs adversarially with a consistent loss. %We also introduce a new method to facilitate an exemplar-based visualization to human users by tracing outputs back to features in Section\ref{sec:interpretation}. 

\subsection{Inconsistency between Feature Representations and Semantic Concepts}
\label{sec:inconsist}

It is a popular statement that deep visual representations have good transferability since they are disentangled~\cite{zhou2014object,zeiler2014visualizing,bau2017network}. 
%, i.e., some neurons can detect semantic concepts spontaneously, and thus form human-interpretable representations\cite{zhou2014object,zeiler2014visualizing,bau2017network}.
In this section, we demonstrate the weakness of this traditional view by showing that the neurons which detect high-level semantics\footnote{In this paper, we roughly mean the high-level visual concepts by objects and parts, while low-level features include colors and textures, and mid-level concepts include attributes (e.g., shiny) and shapes.} (e.g., objects/parts) in DNNs are ambiguous.

\begin{figure*}[!t]
  \centering
  \includegraphics[width=1.0\linewidth]{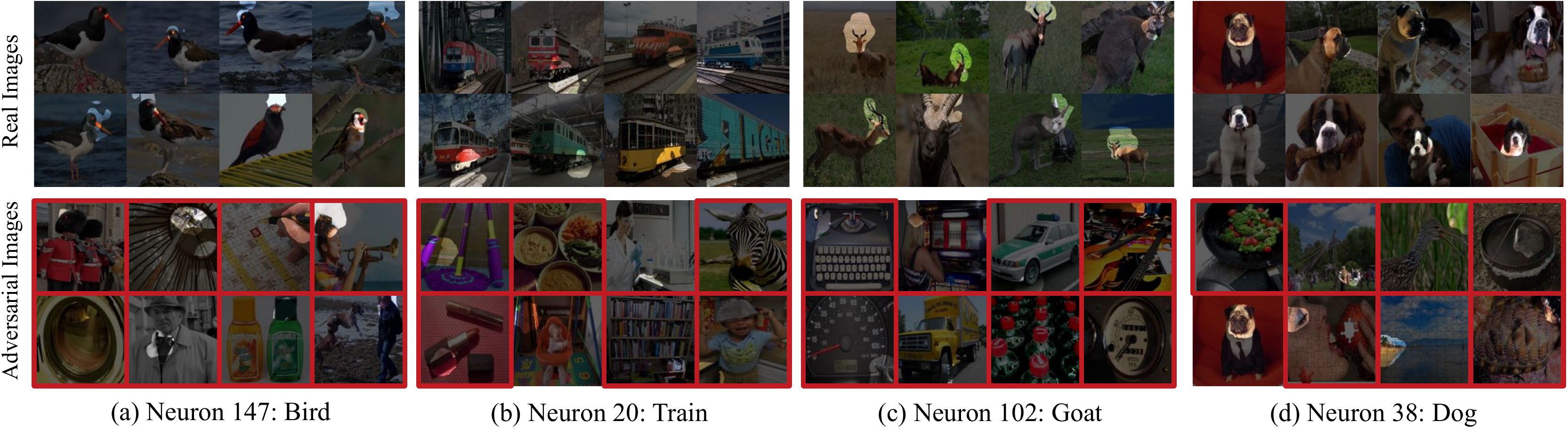}
  \caption{The real and adversarial images with highest activations for neurons in VGG-16 \textit{conv5\_3} layer. }
% It demonstrates that the neurons have explicit semantic meanings when showing real images only, which do not appear in adversarial images.
% Adversarial examples in the red boxes are misclassified to the classes related to the meaning of neurons.
% For example, the \textit{neuron 147} in (a) detects the concept \textit{bird head} in real images, but it fires for adversarial images with various objects. So it is hard to interpret the features learned by neurons in the adversarial setting. This briefly demonstrates the inconsistency of neurons.}
  \label{fig:VGG16-features}
\end{figure*}

% To examine the ``authentic'' internal representations of DNNs when facing adversarial examples, we first use the optimization-based method to generate targeted adversarial examples. 
We first generate a targeted adversarial example $\bm{x}^*$ by solving
\begin{equation}
\label{eq:opt-target}
\argmin_{\bm{x}^*}\|\bm{x}^*-\bm{x}\|_2 + \lambda\cdot\ell(\bm{x}^*, y^*).
\end{equation}
% We therefore generate an adversarial example belonging to the target class $y^*$ when the optimization problem reaches its minimum.
We input these adversarial examples as well as the corresponding real examples into DNNs and examine their internal representations by an activation maximization method~\cite{zhou2014object}. 
%which enables us to visualize and understand what patterns a neuron truly learns to detect.

We show some visualization results in  \autoref{fig:VGG16-features}, where the network is VGG-16 and the images are from the ImageNet dataset~\cite{russakovsky2015imagenet}. The highlighted regions are found by discrepancy map~\cite{zhou2014object}. 
%Specifically, we show $8$ real images and $8$ adversarial images with highest activations for each neuron to represent its learned features. 
% The highlighted regions are found by discrepancy map\cite{zhou2014object}, i.e., the given patch
% is important if there is a large discrepancy and vice versa. Full experimental settings can be found in Section ~\ref{sec:exp}.
As shown in the first row of \autoref{fig:VGG16-features}, the neurons reveal explicit explanatory semantic meanings or human-interpretable concepts when showing real images only, but the contents of the adversarial images do not align with the semantic meanings of the corresponding neurons, if we look at the second row.% The results demonstrate the existence of singular points, which constitute the image subset where neurons are inconsistent with humans' intuition.

\subsection{Adversarial Training with a Consistent Loss}
\label{sec:training}

Based on the above analysis, we propose a novel adversarial training algorithm to train DNNs for the purpose of improving the consistency of neurons. %Unlike other methods that use high-level semantic attributes explicitly\cite{dong2017improving}, we train the DNNs towards interpretability implicitly. Adversarial training has the potential to train interpretable DNNs because it guides the models to learn representations of adversarial images that resemble those of the original images by suppressing the perturbations. Therefore, the neurons are consistently activated when the preferred semantic concepts appear, while deactivated when they disappear, without the interference of adversarial noises.
We introduce a consistent (feature matching) loss in adversarial training. Specifically, we train a DNN parameterized by $\theta$ via minimizing an adversarial objective as
\begin{equation}\label{eq:at-con}
\begin{split}
\min_{\theta} L(\theta),
\quad L(\theta) = \mathbb{E}_{(x,y)\sim\mathcal{D}}\big[\max_{x^*\in \mathcal{S}(x)}\ell(\theta, x^*, y)\\
+ \max_{x'\in \mathcal{S}(x)}d(\phi_{\theta}(x),\phi_{\theta}(x'))\big],
\end{split}
\end{equation}
\newline
where $x$ is the real example; $x^\ast$ and $x'$ correspond to two adversarial examples from the set of all possible adversarial examples $\mathcal{S}(x)$; $\phi_{\theta}$ is the feature representation of interest, and $d$ is a distance metric to quantify the distance between the feature representations of the adversarial example and the corresponding real example. The second term in Eq.~\eqref{eq:at-con} is the consistent loss, which aims to make the feature representation of the worst-case adversarial example close to that of the real example. 
% To solve such a problem, we need to solve the inner maximization problem as well as the outer minimization problem. In particular,
There are two inner maximization problems in Eq.~\eqref{eq:at-con}, so we need to generate two adversarial examples for one real example in principle. But for simplicity, we make the relaxation that we only generate one adversarial example $x^*$ by solving $\max_{x^*\in \mathcal{S}(x)}\ell(\theta, x^*, y)$ and use $x^*$ as an approximation of $x'$ for training. %For finding $x^*$ that solves the inner maximization problem, any of the adversarial sample generation methods introduced in Section~\ref{sec:adv_train} can be used. But from a practical perspective, only the adversarial examples generated by the fast gradient sign method (FGSM) can be used for training DNNs successfully\cite{kurakin2016scale,tramer2017ensemble} on large scale datasets (e.g., ImageNet\cite{russakovsky2015imagenet}), as far as we are concerned.
We choose FGSM to generate adversarial examples, and use them as well as the real examples to train the classifiers by solving the outer minimization problem as
\begin{equation}
\begin{split}
\label{eq:at-train}
\min_{\theta} L(\theta) = \mathbb{E}_{(x,y)\sim\mathcal{D}}\big[\alpha \ell(\theta, x, y) + (1-\alpha)\ell(\theta, x^*, y)\\
+ \beta d(\phi_{\theta}(x),\phi_{\theta}(x^*))\big],
\end{split}
\end{equation}
where $\alpha$, $\beta$ are two balanced weights for these three loss terms. %We use the squared Euclidean distance as the distance metric $d$ and use the last fully-connected layer of each DNN as the feature representation $\phi_{\theta}$, since the last fully-connected layer is most adjacent to the output layer and directly impact the network prediction.

\section{Experiments}
\label{sec:exp}

% In this section, we show the experimental results to validate the effectiveness of our proposed methods. We train three types of DNNs on the ImageNet dataset\cite{russakovsky2015imagenet} and compare their interpretability with normally trained DNNs. We visualize the learned features of neurons as well as compare the performance under our proposed metrics in Section ~\ref{sec:consis_metric}. Experiments demonstrate that the learned features of DNNs trained by our method achieve a satisfactory consistency in both the whole image space and the adversarial image subset.

\subsection{Setup}
\subsubsection{Training}
We use three network architectures---AlexNet~\cite{krizhevsky2012imagenet}, VGG-16~\cite{simonyan2014very} and ResNet-18~\cite{he2015deep} trained on the ImageNet dataset~\cite{russakovsky2015imagenet} in our experiments.
%We use three network architectures---AlexNet\cite{krizhevsky2012imagenet}, VGG-16~\cite{simonyan2014very} and ResNet-18~\cite{he2015deep} trained on the ImageNet dataset~\cite{russakovsky2015imagenet} in our experiments. For normally trained models, We use the pre-trained AlexNet, VGG-16 from Caffe~\cite{jia2014caffe} model zoo\footnote{\url{ https://github.com/BVLC/caffe/wiki/Model-Zoo}} and train the ResNet-18~\cite{he2015deep} model from scratch. The ResNet-18 network is trained with a SGD solver with momentum $0.9$, weight decay $5\times10^{-5}$ and batch size 100. The learning rate starts from $0.05$ and is divided by $10$ when the error plateaus. We train it for $300$K iterations. Due to the computation complexity of adversarial training from scratch~\cite{kurakin2016scale}, we fine-tune the models by Eq.~\eqref{eq:at-train} with $\alpha=0.5$ and $\beta=0.01$. The adversarial examples are generated by FGSM in Eq.~\eqref{eq:fgsm} with the size of perturbations $\epsilon$ ranging from $4$ to $16$. We denote the new models after adversarial training as ``AlexNet-Adv'', ``VGG-16-Adv'' and ``ResNet-18-Adv'', respectively.

\subsubsection{Dataset}
We include two datasets to evaluate the performance. The first dataset is the ImageNet validation set. 
%To examine the consistency of neurons constrained on adversarial examples, we use FGSM (where $\epsilon$ is set to $1$) to generate adversarial samples. We generate non-targeted adversarial images for each real image in the validation set.
For each network, we generate non-targeted adversarial examples using FGSM.
The second dataset is the Broden~\cite{bau2017network} dataset, which provides the densely labeled images of a broad range of visual concepts%, including colors, textures, materials, scenes, parts and objects. This dataset is designed to quantify the interpretability of DNN models. 
 We use iterative least-likely class method to generate adversarial samples in this dataset.% (where $\epsilon$ is set to 8 and the number of iterations is set to 5).

%Given the two datasets as well as the corresponding generated adversarial images, we thereafter qualitatively and quantitatively compare the interpretability of our models with ordinary models.

\subsubsection{Evaluation}
\label{sec:metric}
We use two metrics to evaluate the consistency between the learned features of neurons and semantic concepts.
The first metric is our consistency level proposed in Eq.~\eqref{equ_INPscore}. %For a neuron $n$, we evaluate its consistency level in the whole image space and the adversarial image subset respectively. We denote them as $consis(n)$ and $consis_{adv}(n)$ respectively, following the notations in Section ~\ref{sec:consis_metric}. 
We apply the first metric on the ImageNet validation set.
The second metric is proposed by ~\cite{bau2017network} using the Broden dataset. This metric measures the alignment of neurons with semantic concepts.
%For each neuron/filter, the distribution of activations is computed over the dataset. Then a threshold is determined by the top $0.5\%$ quantile. The neuron activations are masked given the threshold to represent the neuron's semantics. The low-resolution feature masks are further scaled up to the image-resolution segmentation masks. Finally, the intersection-over-union (IoU) score is computed over the masks and the ground-truth segmentation masks of each semantic concept to score the consistency between each neuron and each semantic concept in the dataset.
%However, this metric can be used only given the densely labeled datasets containing the segmentations for various visual concepts. 

\subsection{Experimental Results}

In this section, we compare the interpretability between the models obtained by our proposed adversarial training algorithm with those trained normally.

\begin{figure*}[t]
  \centering
  \includegraphics[width=1.0\linewidth]{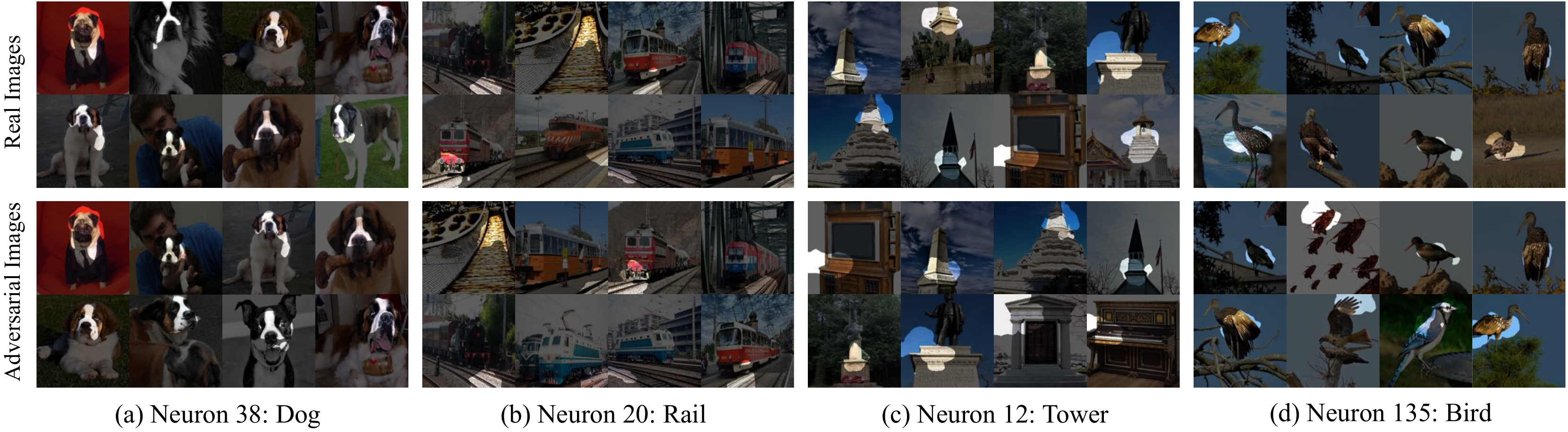}
  \caption{The real and adversarial images with highest activations for neurons in VGG-16-Adv \textit{conv5\_3} layer.}
  %The model is obtained by adversarial training with the consistent loss. The visual concepts in both of them are quite similar. The results prove that we eliminate singular points, which originally exist in the adversarial image subset.}
  \label{fig:VGG16-features-adv}
\end{figure*}

\subsubsection{Experimental Results on ImageNet Dataset}
We first show our visualization results. %We follow the activation maximization method proposed in~\cite{zhou2014object} to visualize the learned features of each neuron. 
The results are shown in \autoref{fig:VGG16-features} and \autoref{fig:VGG16-features-adv} for neurons in VGG-16 and VGG-16-Adv, respectively. 
%For each neuron, we show 8 real images and 8 adversarial images from the ImageNet validation set with highest activations to represent its learned features. As we already discussed in Section~\ref{sec:inconsist}, the neurons do not respond to high-level semantic concepts when showing adversarial images.
We find that, after adversarial training, the visual concepts are quite similar in both of the real images and adversarial images.%, indicating that we do eliminate these singular points which originally exist in adversarial image subset. 
This result shows that the network trained with a consistent loss is more interpretable than the normally trained network. 

\begin{figure}[t]
  \centering
  \includegraphics[width=1.0\linewidth]{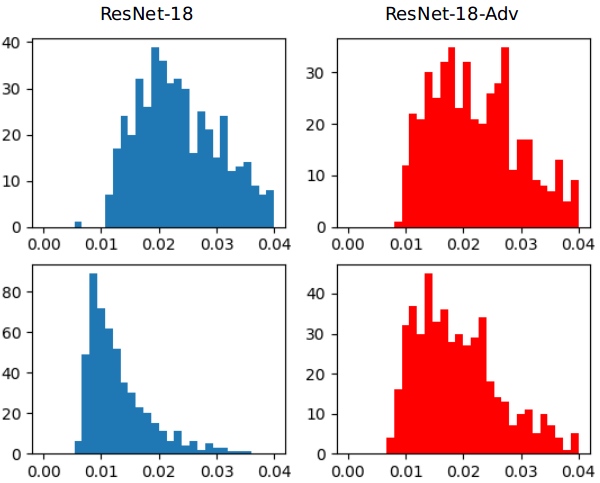}
  \caption{The distribution of $consis(n)$ and $consis_{adv}(n)$ of neurons in the last convolutional layer of ResNet-18 and ResNet-18-Adv. The left two histograms corresponds to ResNet-18 and the right two histograms corresponds to ResNet-18-Adv. The top two histograms corresponds to $consis(n)$ and the bottom two histograms corresponds to $consis_{adv}(n)$.}
  \label{fig:consis_dis}
\end{figure}

Then we show our quantitave results.
%We use our consistency level proposed in Section ~\ref{sec:consis_metric} to evaluate the performance of networks.
We show the average of $consis(n)$ and $consis_{adv}(n)$ of neurons in the last convolutional layer.
As shown in \autoref{tab:consis_last_conv}, the consistency level of neurons on the whole image space don't change much after adversarial training, while the consistency level constrained on the adversarial image subset increases. 
%This result indicates that the adversarial training with a consistent loss successfully eliminate these singular points which originally exist in the adversarial image subset, without harming the consistency level on the whole image space. 
We also plot the distribution of $consis(n)$ and $consis_{adv}(n)$ of neurons in the last convolutional layer of ResNet-18 and ResNet-18-Adv. As shown in ~\autoref{fig:consis_dis}, when showing adversarial samples, the distribution shift left sharply for ResNet-18, while ResNet-18-Adv doesn't change much.
%These evidences all prove that our adversarial training algorithm with a consistent loss can help a network learn more consistent features on adversarial image subset, meanwhile preserving the consistency on the whole image space. Besides, we also evaluate the consistency level on lower layers. For lower layers, they share similar distribution of $consis(n)$ regardless of how the network is trained and on which dataset the network is evaluated, as shown in ~\autoref{fig:consis_dis_lower}. It is because the neurons in lower layers detect low-level concepts, such as texture and colors. These low-level concepts don't have corresponding labels in ImageNet dataset and are shared by different high-level concepts. So neurons in lower layers activate images with different high-level concepts, causing their consistency level lower than normal. To evaluate concepts of different semantic level, we use Broden dataset, which integrates many low-level concepts. This dataset allows us to evaluate neurons regardless of their semantic levels. 

\begin{table}[!t]
\begin{center}
\caption{The average of $consis(n)$ and $consis_{adv}(n)$ of neurons in the last convolutional layer in each network. The $consis(n)$ is calculated on the whole image space and the $consis_{adv}(n)$ is calculated on the adversarial image subset.}
\vspace{2ex}
\begin{tabular}{c|c|c}
\hline\hline
&$consis(n)$&$consis_{adv}(n)$\\
\hline\hline
AlexNet & 0.0162 & 0.0116 \\
AlexNet-Adv & 0.0161 & 0.0121\\
\hline
VGG-16 & 0.0296 & 0.0187 \\
VGG-16-Adv & 0.0287 & 0.0220\\
\hline
ResNet-18 & 0.0261 & 0.0133\\
ResNet-18-Adv & 0.0251 & 0.0206\\
\hline
\end{tabular}
\label{tab:consis_last_conv}
\end{center}
\end{table}

% \begin{figure}[t]
%   \centering
%   \includegraphics[width=1.0\linewidth]{images/consis_dis_lower.png}
%   \caption{The distribution of $consis(n)$ and $consis_{adv}(n)$ of neurons in a lower layer(relu3b) of ResNet-18 and ResNet-18-Adv. The left two histograms corresponds to ResNet-18 and the right two histograms corresponds to ResNet-18-Adv. The top two histograms corresponds to $consis(n)$ and the bottom two histograms corresponds to $consis_{adv}(n)$.}
%   \label{fig:consis_dis_lower}
% \end{figure}

\subsubsection{Experimental Results on Broden Dataset}

\begin{table}[!t]
\begin{center}
\caption{The ratio (\%) of neurons that align with semantic concepts for each model, when showing real and adversarial images respectively. There are six types of concepts: colors (C.), textures (T.), materials (M.), scenes (S.), parts (P.) and objects (O.).}
\vspace{2ex}
\begin{tabular}{c|p{3.0ex}<{\centering}p{3.0ex}<{\centering}p{3.0ex}<{\centering}p{3.0ex}<{\centering}p{3.0ex}<{\centering}p{3.0ex}<{\centering}}
\hline\hline
\multirow{2}{*}{} & \multicolumn{6}{c}{Real Images} \\
\cline{2-7}
& C. & T. & M. & S. & P. & O.\\
\hline\hline
AlexNet & 0.4 & 13.3 & 0.8 & 0.4 & 6.6 & 17.2\\
AlexNet-Adv & 0.4 & 20.7 & 0.0 & 0.4 & 4.3 & 16.4\\
\hline
VGG-16 & 0.2 & 11.9 & 0.8 & 4.7 & 10.2 & 34.6\\
VGG-16-Adv & 0.2 & 18.0 & 1.2 & 5.7 & 8.6 & 30.9\\
\hline
ResNet-18 & 0.0 & 12.7 & 0.8 & 5.9 & 4.3 & 33.8\\
ResNet-18-Adv & 0.2 & 21.3 & 1.6 & 6.1 & 6.8 & 33.8\\
\end{tabular}
\begin{tabular}{c|p{3.0ex}<{\centering}p{3.0ex}<{\centering}p{3.0ex}<{\centering}p{3.0ex}<{\centering}p{3.0ex}<{\centering}p{3.0ex}<{\centering}}
\hline\hline
\multirow{2}{*}{} & \multicolumn{6}{c}{Adversarial Images} \\
\cline{2-7}
& C. & T. & M. & S. & P. & O.\\
\hline\hline
AlexNet & 0.4 & 12.1 & 0.4 & 0.0 & 5.5 & 10.2\\
AlexNet-Adv & 0.4 & 18.8 & 0.0 & 0.4 & 4.3 & 11.3\\
\hline
VGG-16 & 0.2 & 2.9 & 0.0 & 0.2 & 3.3 & 6.4\\
VGG-16-Adv & 0.4 & 14.1 & 0.6 & 1.4 & 6.4 & 17.4\\
\hline
ResNet-18 & 0.0 & 4.3 & 0.2 & 0.6 & 1.0 & 6.1\\
ResNet-18-Adv & 0.2 & 15.0 & 0.4 & 1.2 & 3.1 & 6.8\\
\hline
\end{tabular}

\label{tab:dissect}
\end{center}
\end{table}

For the Broden dataset, we calculate the number of neurons that align with visual concepts for each model. 
%We consider one neuron as a detector for a semantic concept if the IoU score between the neuron and the concept is larger than $0.04$. 
For each model, we also generate a set of adversarial images to evaluate the interpretability of neurons. We show the results in \autoref{tab:dissect}.
It can be seen that for normally trained models, the number of neurons that align with both high-level and low-level semantic concepts decreases significantly by showing adversarial images. 
%It proves that the learned features of neurons are ambiguous in the presence of adversarial examples. 
On the other hand, the neurons in adversarially trained models are more consistent, in the presence of adversarial images. % The results prove that the models trained by the proposed adversarial training algorithm learn more consistent feature representations in the presence of adversarial examples.

\subsubsection{Model Performance}

\begin{table}[t]
\begin{center}
\caption{Accuracy (\%) on the ImageNet validation set and adversarial examples generated by FGSM with $\epsilon=1$.}
\begin{tabular}{c|c|p{8.0ex}<{\centering}|p{8.0ex}<{\centering}|p{8.0ex}<{\centering}}
\hline
& & AlexNet  & VGG16  & ResNet18 \\
\hline
\multirow{2}{*}{Real} & Top-1 & 54.58 & 68.15 & 66.30\\
\cline{2-5}
& Top-5 & 78.17 & 88.30 & 87.09\\
\hline
\multirow{2}{*}{Adv.} & Top-1 & 4.44 & 8.60 & 4.41\\
\cline{2-5}
& Top-5 & 22.94 & 36.94 & 31.8\\
\hline\hline
& & AlexNet-Adv & VGG16-Adv & ResNet18-Adv \\
\hline
\multirow{2}{*}{Real} & Top-1 & 43.92 & 62.55 & 54.01\\
\cline{2-5}
& Top-5 & 68.80 & 84.66 & 77.84\\
\hline
\multirow{2}{*}{Adv.} & Top-1 & 17.45 & 25.62 & 27.56\\
\cline{2-5}
& Top-5 & 38.12 & 56.17 & 55.61\\
\hline
\end{tabular}

\label{tab:performance}
\end{center}
\end{table}

We report the performance of these models on the ImageNet validation set as well as the adversarial examples generated by FGSM in \autoref{tab:performance}. We notice that after adversarial training, the accuracy of the models drops about $10\%$. But the models after adversarial training can improve the accuracy against attacks. %We argue that adversarial training is a trade-off between models' performance and their interpretability as well as the robustness~\cite{Madry2017Towards,DoshiKim2017Interpretability}.

\bibliography{refs}

\begin{thebibliography}{}

\bibitem[\protect\citeauthoryear{Bau \bgroup et al\mbox.\egroup
  }{2017}]{bau2017network}
Bau, D.; Zhou, B.; Khosla, A.; Oliva, A.; and Torralba, A.
\newblock 2017.
\newblock Network dissection: Quantifying interpretability of deep visual
  representations.
\newblock In {\em CVPR}.

\bibitem[\protect\citeauthoryear{Doshi-Velez}{2017}]{DoshiKim2017Interpretability}
Doshi-Velez, Finale;~Kim, B.
\newblock 2017.
\newblock Towards a rigorous science of interpretable machine learning.
\newblock In {\em eprint arXiv:1702.08608}.

\bibitem[\protect\citeauthoryear{He \bgroup et al\mbox.\egroup
  }{2016}]{he2015deep}
He, K.; Zhang, X.; Ren, S.; and Sun, J.
\newblock 2016.
\newblock Deep residual learning for image recognition.
\newblock In {\em CVPR}.

\bibitem[\protect\citeauthoryear{Krizhevsky, Sutskever, and
  Hinton}{2012}]{krizhevsky2012imagenet}
Krizhevsky, A.; Sutskever, I.; and Hinton, G.~E.
\newblock 2012.
\newblock Imagenet classification with deep convolutional neural networks.
\newblock In {\em NIPS}.

\bibitem[\protect\citeauthoryear{Miller \bgroup et al\mbox.\egroup
  }{1990}]{miller1990introduction}
Miller, G.~A.; Beckwith, R.; Fellbaum, C.; Gross, D.; and Miller, K.~J.
\newblock 1990.
\newblock Introduction to wordnet: An on-line lexical database.
\newblock {\em International journal of lexicography} 3(4):235--244.

\bibitem[\protect\citeauthoryear{Russakovsky \bgroup et al\mbox.\egroup
  }{2015}]{russakovsky2015imagenet}
Russakovsky, O.; Deng, J.; Su, H.; Krause, J.; Satheesh, S.; Ma, S.; Huang, Z.;
  Karpathy, A.; Khosla, A.; Bernstein, M.; et~al.
\newblock 2015.
\newblock Imagenet large scale visual recognition challenge.
\newblock {\em International Journal of Computer Vision} 115(3):211--252.

\bibitem[\protect\citeauthoryear{Simonyan and
  Zisserman}{2015}]{simonyan2014very}
Simonyan, K., and Zisserman, A.
\newblock 2015.
\newblock Very deep convolutional networks for large-scale image recognition.
\newblock In {\em ICLR}.

\bibitem[\protect\citeauthoryear{Zeiler and
  Fergus}{2014}]{zeiler2014visualizing}
Zeiler, M.~D., and Fergus, R.
\newblock 2014.
\newblock Visualizing and understanding convolutional networks.
\newblock In {\em ECCV}.

\bibitem[\protect\citeauthoryear{Zhou \bgroup et al\mbox.\egroup
  }{2015}]{zhou2014object}
Zhou, B.; Khosla, A.; Lapedriza, A.; Oliva, A.; and Torralba, A.
\newblock 2015.
\newblock Object detectors emerge in deep scene cnns.
\newblock In {\em ICLR}.

\end{thebibliography}
\bibliographystyle{aaai}

\end{document}